\pdfoutput=1
\documentclass[11pt]{article}
\usepackage{acl}
\usepackage{times}
\usepackage{latexsym}
\usepackage[T1]{fontenc}
\usepackage[utf8]{inputenc}
\usepackage{microtype}
\usepackage{enumitem}
\usepackage{booktabs}
\usepackage{xspace}
\newcommand{\anli}{$\alpha$NLI\xspace}
\usepackage{amsmath}
\DeclareMathOperator*{\argmax}{arg\,max}
\usepackage{pgfplots}

\title{Embarrassingly Simple Performance Prediction for\\ Abductive Natural Language Inference}

\author{Emīls Kadiķis, Vaibhav Srivastav, \and Roman Klinger \\
  Institut für Maschinelle Sprachverarbeitung \\
  University of Stuttgart \\
  Pfaffenwaldring 5b, 70569 Stuttgart \\
  \texttt{\{emils.kadikis,vaibhav.srivastav,klinger@ims.uni-stuttgart.de\}}
  }

\begin{document}
\maketitle

\begin{abstract}
  The task of abductive natural language inference (\anli), to decide
  which hypothesis is the more likely explanation for a set of
  observations, is a particularly difficult type of NLI.  Instead of
  just determining a causal relationship, it requires common sense to
  also evaluate how reasonable an explanation is.  All recent
  competitive systems build on top of contextualized representations
  and make use of transformer architectures for learning an NLI
  model. When somebody is faced with a particular NLI task, they need
  to select the best model that is available. This is a time-consuming
  and resource-intense endeavour.  To solve this practical problem, we
  propose a simple method for predicting the performance without
  actually fine-tuning the model. We do this by testing how well the
  pre-trained models perform on the \anli task when just comparing
  sentence embeddings with cosine similarity to what the performance
  that is achieved when training a classifier on top of these
  embeddings. We show that the accuracy of the cosine similarity
  approach correlates strongly with the accuracy of the classification
  approach with a Pearson correlation coefficient of 0.65. Since the
  similarity computation is orders of magnitude faster to compute on a
  given dataset (less than a minute vs.\ hours), our method can lead
  to significant time savings in the process of model selection.
\end{abstract}

\section{Introduction}
Abduction is a type of reasoning that infers an explanation for some
observations \citep{peirce1931collected}. It is a particularly
challenging type of inference; as opposed to deduction and induction,
which derive the conclusion from only the information present in the
observations, abduction requires making assumptions about an implicit
context beyond just the given observations. Abductive reasoning is
therefore at the core of the way humans understand the world and how
world knowledge is involved.

Abductive reasoning in the natural language domain has been introduced
by \citet{Bhagavatula2020} who defined the abductive natural
language inference (\anli) task. In it, we are given four sentences --
two observations $o_1$ and $o_2$ and two hypotheses $h_1$ and $h_2$,
where we know that the sequence of events was
$o_1 \rightarrow (h_1|h_2) \rightarrow o_2$.  The task then is to
decide which of the two hypotheses is the more plausible one.

An example from \citet{Bhagavatula2020} is the following:
\begin{description}[noitemsep]
\item[$o_1$]: It was lunchtime and Kat was hungry.
\item[$o_2$]: Kat and her coworkers enjoyed a nice lunch outside.
\item[$h_1$]: Kat went to get a salad.
\item[$h_2$]: Kat decided to take a nap instead of eating.
\end{description}
While it is not inconceivable that someone would decide to take a nap on their lunch
break ($h_2$), given $o_2$ the first hypothesis becomes more likely.

Currently, transformer-based architectures
\citep{DBLP:journals/corr/VaswaniSPUJGKP17} are state of the art in a
wide variety of natural language processing (NLP) tasks
\citep{devlin2019bert, he2021deberta, li2020unimo}, including
\anli. However, with an ever-changing landscape of transformer models
and pre-training techniques (with over
10000\footnote{https://huggingface.co/models} different fine-tuned
models available on the HuggingFace hub
\citep{wolf-etal-2020-transformers}), finding the best model for a
given task has become a time-consuming process since, in order to
compare multiple models, they each need to be separately fine-tuned on
the task.

This model selection process might lead to a prohibitive runtime,
which has led to research on performance prediction, namely to predict
the expected performance out of parameters of the model configuration,
without actually training the model. This procedure has been evaluated
for a set of NLP tasks, including span prediction
\citep{papay-etal-2020-dissecting} and language modelling
\citep{chen-2009-performance}. However, we are not aware of any
previous work that performed performance prediction for \anli.

In this paper, we introduce a fast performance prediction method for
the \anli task that allows a more guided way of choosing which models
to fine-tune.  We use various pre-trained transformer models to embed
the observations and hypotheses with the approach introduced in
\citet{reimers-2019-sentence-bert}, then compare which hypothesis is
closer to the observations with cosine similarity.  We find that the
performance of the similarity-based approach is correlated to results
obtained via fine-tuning. Therefore, the similarity-based approach can
serve as a performance prediction method.

\section{Related work}
There are three research topics that need to be mentioned. Approaches
to abductive reasoning, pre-trained language models, and performance
prediction. In this section, we explore them in detail.

\paragraph{Abductive natural language inference.}
NLI has been proposed as the task of recognizing textual entailment by
\citet{maccartney-manning-2008-modeling} and now constitutes a major
challenge in NLP which has found application for other downstream tasks,
including question answering or zero-shot classification
\citep{Yin2019,Mishra2020}. Based on the initial goal of establishing
inference relations between two short texts, a myriad of variants have
been proposed \citep{yin-etal-2021-docnli, williams-etal-2018-broad,
  bowman-etal-2015-large}.  One such variant is abductive NLI
\citep[\anli,][]{Bhagavatula2020}.

Transformer-based architectures dominate the \anli
leaderboard,\footnote{\url{https://leaderboard.allenai.org/anli/submissions/public}}
including RoBERTa-based models
\citep{liu2020roberta,mitra2020additional} which explore how
additional knowledge can improve performance on reasoning tasks, and
\citet{Zhu2020} who approach \anli as a ranking task.  The task
authors improved upon their result in \citet{lourie2021unicorn} by
using a T5 model \citep{2020t5} and experimenting with multi-task
training and fine-tuning over multiple reasoning tasks. Both the
multi-task criteria and the Text-to-Text framework of T5 help the
model generalize and understand the context better.

The second-best model on the leaderboard is a DeBERTa model
\citep{he2021deberta}. The model replaces the masked language
modelling task with a replaced-token detection task. It also uses a
disentangled attention mechanism to encode the position and content
information.

The current state of the art shows an accuracy of 91.18\% using a new
unified-modal pre-training method to leverage multi-modal data for
single-modal tasks \citep{li2020unimo}. This result approaches the
human baseline of 92.9\%.

\paragraph{Pre-trained language models.}
The \anli task requires the model to successfully ``understand'' the
context of both the observations and use that understanding to
identify the more likely hypothesis entailing it. Most semantic
representations in practical applications rely on distributional
semantics. Such methods include the word-level embedding methods
Word2Vec \citep{DBLP:journals/corr/abs-1301-3781} and GloVe
\citep{pennington2014glove} and language model-based word
representation like ELMo \citep{Peters:2018}, ULMFit
\citep{howard-ruder-2018-universal}, and GPT
\cite{radford2018improving}.

The current state of the art are pre-trained transformer architectures
\citep{DBLP:journals/corr/VaswaniSPUJGKP17} like BERT \citep{devlin2019bert}, which use a masked language modelling
and next sentence prediction objective. This not only helps the model
understand the context within a sentence but also in-between
consecutive sentences. There is however a trade-off in terms of the
time it takes to train transformer models. For example, a from-scratch
training of BERT takes 6.4 days on an 8 GPU Nvidia V100
server\footnote{https://aws.amazon.com/blogs/machine-learning/amazon-web-services-achieves-fastest-training-times-for-bert-and-mask-r-cnn/}. 
\citet{devlin2019bert} recommend fine-tuning the language
model between 2-4 epochs for a given task. However, in practice,
multiple trials are required to find the optimal
hyperparameters. These long training times and multiple fine-tuning
runs make model selection a time-intensive process
\citep{liu-wang-2021-empirical}.

\paragraph{Performance prediction.} The task of performance prediction
is to estimate the performance of a specific system without explicitly
running it. It helps in setting hyperparameters, finding
feature sets, or identifying candidate language models for a
downstream NLP task.  \citet{chen-2009-performance} develop, for
instance, a generalized method for predicting the performance of
exponential language models. They analyze the backoff features in an
exponential $n$-gram model.  \citet{papay-etal-2020-dissecting}
leverage meta-learning to identify candidate model performance on the
task of span identification. They train a linear regressor as a
meta-model to predict span ID performance based on model features and
task metrics for an unseen task.  \citet{ye-etal-2021-towards} propose
performance prediction methods particularly suited for fine-grained
evaluation metrics. They also develop methods for estimating the
reliability of these performance predictions.
Contrary to the previously mentioned papers,
\citet{xia-etal-2020-predicting} build regression models to predict
the performance across a variety of NLP tasks, however, they do not
consider NLI as one of them.

In contrast to our work, all these previous methods build on top of
the idea to train a surrogate model for performance prediction and
depend on the information about past runs of these models. Our
approach focuses solely on the embeddings provided by the language
model and leverages those as a predictor of performance. This particular setup is also motivated by the \anli task itself, in
which a sentence needs to be chosen for a given set of other
sentences.

\section{Methods}
Our paper investigates how well a fine-tuned transformer model's
performance on the \anli task \citep{Bhagavatula2020} is approximated
by the cosine similarity of embeddings of the input sentences which we
obtain from the pre-trained models before fine-tuning them.

Intuitively, if a model embeds the correct hypothesis close to the
observations in some latent space (not necessarily a semantic
similarity space), then a classification model built on top of that
latent space should have an easier time discerning which is the
correct hypothesis, because apparently that latent space captured some
features that were salient for the \anli task.

\subsection{Sentence Embedding}
For both the similarity baseline and the fine-tuned classification
model, the starting point is the pre-trained transformer model
itself. We add a mean pooling layer to convert the token-by-token
output of the model into a single sentence embedding
\citep{reimers-2019-sentence-bert}.  Given some tokenized input
$\mathbf{x}=[x_1,x_2,\ldots,x_n]$ and a pre-trained transformer model
$E$ which encodes each token $E(x_i)$, we calculate the sentence
embedding $\mathrm{emb}(\mathbf{x})=\frac{1}{n} \sum_{i=1}^{n}E(x_i)$.

Al alternative to mean pooling would have been to use the embedding of
the CLS token. We opted against that for three reasons. Firstly,
\citet{reimers-2019-sentence-bert} show that mean pooling slightly
outperforms using the CLS token in their semantic similarity models.
Secondly, for some models, the CLS token does not have any particular
significance before fine-tuning on the downstream task due to the
training objective they use (such as RoBERTa \cite{liu2020roberta},
which uses masked language modelling).  Thirdly, pooling is a general
approach that can be adopted for any model, even if it does not output
a CLS token. Since our goal was to avoid any model-specific
enhancements, a universal blanket approach like this was preferable.

\subsection{Similarity-based \anli}

To perform \anli on a validation instance, we obtain three
sentence embeddings -- one for the combined observations $o_1 + o_2$
and one for each of the hypotheses $h_1$, $h_2$.
To predict the more plausible hypothesis, we calculate which of them
is closer to the observations with cosine similarity:
\[
  \hat{h} = \argmax_{h'\in \{h_1,h_2\}}\cos(\mathrm{emb}(o_1 + o_2),\mathrm{emb}(h'))
\]

\subsection{Classification-based \anli}
For the classification model, we add a classification head on top of
the pre-trained model, which consists of a mean pooling layer to get
sentence embeddings, then a fully-connected layer and a softmax output
layer.  For each instance of ($o_1, o_2, h_1, h_2$), the model takes
two different inputs which consist of both observations with each of
the hypotheses, namely $\mathrm{emb}(o_1 + o_2 + h_1)$
and $\mathrm{emb} (o_1 + o_2 + h_2)$.

Both of these input representations are then used in a fully connected
layer $f$ with a softmax output layer to get the probability score for
each input.  The hypothesis that is assigned the largest probability
constitutes the prediction:
\[
  \hat{h}' = \argmax_{h\in \{h_1,h_2\}}\mathrm{softmax}(f(\mathrm{emb}(o_1 + o_2 + h)))
\]

In our experiments, we fine-tune the classification head on the \anli
training set without updating weights in the underlying language
model.  This is mostly due to time and resource constraints, however,
we believe that while fine-tuning would improve classification
performance across the board, it would not affect the ranking as
such. Since we are comparing models amongst themselves, the ranking is
more important.

\begin{table*}
  \centering\small
  \setlength{\tabcolsep}{13pt}
\begin{tabular}{llcc cc}
\toprule
  && \multicolumn{2}{c}{Accuracy} & \multicolumn{2}{c}{Run time} \\
  \cmidrule(r){3-4}\cmidrule(l){5-6}
Model & Citation & Sim. & Class. & Sim. & Class.\\
\cmidrule(r){1-1}\cmidrule(rl){2-2}\cmidrule(rl){3-4}\cmidrule(l){5-6}
 albert-base-v2 & Lan (\citeyear{Lan2020}) & 50.78\% & 65.34\% & 5.68 & 0:55:58 \\
 albert-large-v2 & Lan (\citeyear{Lan2020}) & 50.13\% & 69.71\% & 7.93 & 2:51:01 \\
 bert-base-uncased & Devlin (\citeyear{devlin2019bert}) & 51.69\% & 65.99\% & 2.69 & 1:13:28 \\
 bert-large-uncased & Devlin (\citeyear{devlin2019bert}) & 52.67\% & 67.04\% & 7.22 & 3:13:54 \\
 distilbert-base-uncased & Sanh (\citeyear{Sanh2019DistilBERTAD}) & 51.63\% & 62.60\% & 1.63 & 0:26:36 \\
 squeezebert/squeezebert-uncased & Iandola (\citeyear{iandola-etal-2020-squeezebert}) & 50.52\% & 61.95\% & 2.23 & 0:37:00 \\ 
 google/mobilebert-uncased & Sun (\citeyear{sun-etal-2020-mobilebert}) & 48.75\% & 61.68\% & 4.29 & 0:36:20 \\
 google/canine-s & Clark (\citeyear{DBLP:journals/corr/abs-2103-06874}) & 49.21\% & 58.09\% & 3.82 & 1:30:40 \\
 google/electra-small-discriminator & Clark (\citeyear{Clark2020ELECTRA:}) & 51.17\% & 63.51\% & 1.69 & 0:12:27 \\
 google/electra-base-discriminator & Clark (\citeyear{Clark2020ELECTRA:}) & 52.28\% & 78.07\% & 2.77 & 0:51:51 \\
 google/electra-large-discriminator & Clark (\citeyear{Clark2020ELECTRA:}) & 52.74\% & 88.51\% & 7.23 & 3:09:35 \\
 microsoft/mpnet-base & Song (\citeyear{song2020mpnet}) & 51.50\% & 74.87\% & 2.77 & 0:52:36 \\
 roberta-base & Liu (\citeyear{liu2020roberta}) & 51.50\% & 74.15\% & 2.70 & 0:52:28 \\
 roberta-large & Liu (\citeyear{liu2020roberta}) & 51.63\% & 84.14\% & 6.87 & 3:32:47 \\
 google/bigbird-roberta-base &Zaheer (\citeyear{zaheer2020bigbird}) & 51.50\% & 71.02\% & 5.74 & 1:02:08 \\
 kssteven/ibert-roberta-base & Kim (\citeyear{kim2021bert}) & 51.50\% & 73.63\% & 2.78 & 1:00:39 \\
 distilroberta-base & Sanh (\citeyear{Sanh2019DistilBERTAD}) & 51.43\% & 65.99\% & 1.67 & 0:29:08 \\
\bottomrule
\end{tabular}
\caption{Accuracy on the \anli validation set using similarity and a classification model. The similarity runtime (how long it takes to evaluate the model since no training is required) is shown in seconds, the classification runtime (how long it takes to fine-tune and evaluate the model) in hours, minutes, and seconds. Note that the given training time is for a single set of hyperparameters. Identifying the best hyperparameters involved training each model multiple times.}
\label{table:results}
\end{table*}

\section{Experiments}
We compare the similarity-prediction-based \anli approach and the
classification-based \anli approach to evaluate if the first can act
as an approximation for the performance expected by the second.  We
use the pre-trained transformer models which are available on the
HuggingFace \citep{wolf-etal-2020-transformers} hub. The full list of
models we use is listed in Table~\ref{table:results}. The code for the
experiments is available
online.\footnote{https://github.com/Vaibhavs10/anli-performance-prediction}.

\subsection{Dataset}
All of our experiments were run on the train and validation split in
the ART dataset provided for the
\anli challenge \citep{Bhagavatula2020}. It consists of 169,654
training and 1,532 validation samples, each consisting of two
observations and two hypotheses obtained from a narrative short story
corpus and augmented with wrong hypotheses written by crowd-sourced workers.

The training data contains repetitions of the same $(o_1, o_2)$ pairs
with different sets of hypotheses, ranging from one plausible and one
implausible hypothesis to two plausible hypotheses where one of them
is more plausible. The validation set was constructed using
adversarial filtering, which selects one plausible and one implausible hypothesis for each set of observations that are hard to distinguish. This increases the
probability that the instances are free of annotation artifacts, which
the authors defined as ``unintentional patterns in the data that leak
information about the target label'' \citep{Bhagavatula2020}.

\subsection{Experimental Setup}
All of our classification and similarity experiments were run on an
Nvidia RTX 2080 GPU.  For training the classifier we used the maximum
batch size that fit on the GPU (which is different for different sized
models, ranging between 12 and 128). For similarity experiments, we
only infer the embeddings from pre-trained models. For hyperparameter
selection, to keep the comparison fair, we tuned the batch size
and learning rate and considered the same set of possible combinations
across all the models. The specific values used for each model are
available in Table~\ref{table:hyperparameters} in the appendix. We
train for 3 epochs with learning rates ranging
$[10^{-5};9\cdot10^{-5}]$ and a weight decay of 0.01. For each
pre-trained model, we pick the one that achieved the highest accuracy
on the validation set.

\begin{figure}[t]
  \centering\sffamily
  \pgfplotsset{every tick label/.append style={font=\footnotesize},
    label style={font=\small}
  }
  \begin{tikzpicture}
    \begin{axis}[
      height=6cm,
      width=7cm,
      grid=major,
      xlabel={similarity acc.},
      ylabel={classification acc.},
      xmin=45,
    xmax=55,
    ymin=50,
    ymax=100,
      ]
      \addplot[only marks] coordinates {
        (50.78,65.34)
        (50.13,69.71)
        (51.69,65.99)
        (52.67,67.04)
        (51.63,62.60)
        (50.52,61.95)
        (48.75,61.68)
        (49.21,58.09)
        (51.17,63.51)
        (52.28,78.07)
        (52.74,88.51)
        (51.50,74.87)
        (51.50,74.15)
        (51.63,84.14)
        (51.50,71.02)
        (51.50,73.63)
        (51.43,65.99)
      };
    \end{axis}
  \end{tikzpicture}
  \caption{Relation between similarity and classification accuracy
    scores. Note that the similarities are closer to each other than
    the classification values, therefore we chose a different scale.}
  \label{fig:scatter}
\end{figure}
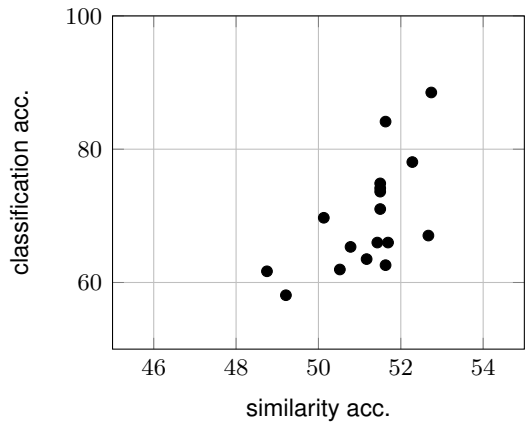

\subsection{Evaluation and Results}

Table~\ref{table:results} shows the results as accuracy scores,
obtained with each pre-trained model when using cosine similarity and
when using a classifier built on top of it. We also show the training
runtimes.

The primary observation is that the accuracy scores of classification
and similarity are correlated. This can be seen in Figure~\ref{fig:scatter}. The Pearson correlation coefficient is
$r=.65$ ($p=0.005$). The Spearman's correlation coefficient is
$\rho=.67$ ($p=0.003$), indicating that the ranking obtained with the
similarity-based prediction is a reliable indicator that is helpful
for model selection. Additionally, model fine-tuning takes on average 620
times longer than the similarity-based estimate. Tuning the
hyperparameters involved training each model multiple times.

\section{Conclusions \& Future Work}
In this paper, we have shown that similarity measures based on
the distributional semantic representation in pre-trained transformer models
serve as an effective proxy for fine-tuned transformer-based classification in
\anli.  Since fine-tuning a transformer model takes notably more time
than performing similarity comparisons, our approach supports
efficient model selection procedures. Future work should investigate
the suitability of similarity-based performance prediction for other
similar tasks, like next sentence prediction, question answering,
summarization.

\section*{Acknowledgements}
We thank the anonymous reviewers and the action editor at ACL Rolling
Review for their helpful comments. This work has been conducted in the
context of projects funded by the
German Research Council (DFG), project ``Computational Event Analysis
based on Appraisal Theories for Emotion Analysis'' (CEAT, project
number KL 2869/1-2) and project ``Automatic Fact Checking for
Biomedical Information in Social Media and Scientific Literature''
(FIBISS, KL 2869/5-1).

\bibliography{custom}
\bibliographystyle{acl_natbib}

\appendix
\section{Model Hyperparameters}
All models were trained for 3 epochs with a weight decay of 0.01. The batch size and learning rate used for each model can be seen in Table~\ref{table:hyperparameters}
\begin{table}[h]
  \small
  \setlength{\tabcolsep}{4pt}
\begin{tabular}{lrr}
\toprule
Model & Learning rate & Batch size\\
\cmidrule(r){1-1}\cmidrule(rl){2-2}\cmidrule(l){3-3}
 albert-base-v2 & $10^{-5}$ & 60 \\
 albert-large-v2 & $10^{-5}$ & 20 \\
 bert-base-uncased & $5\cdot10^{-5}$ & 32 \\
 bert-large-uncased & $10^{-5}$ & 16 \\
 distilbert-base-unc. & $9\cdot10^{-5}$ & 128 \\
 squeezebert/squeezebert-unc. & $7\cdot10^{-5}$ & 64 \\
 google/mobilebert-unc. & $7\cdot10^{-5}$ & 100 \\
 google/canine-s & $10^{-5}$ & 24 \\
 google/electra-small-discr. & $7\cdot10^{-5}$ & 128 \\
 google/electra-base-discr. & $3\cdot10^{-5}$ & 64 \\
 google/electra-large-discr. & $10^{-5}$ & 16 \\
 microsoft/mpnet-base & $3\cdot10^{-5}$ & 64 \\
 roberta-base & $3\cdot10^{-5}$ & 60 \\
 roberta-large & $10^{-5}$ & 12 \\
 google/bigbird-roberta-base & $10^{-5}$ & 40 \\
 kssteven/ibert-roberta-base & $3\cdot10^{-5}$ & 64 \\
 distilroberta-base & $5\cdot10^{-5}$ & 80 \\
\bottomrule
\end{tabular}
\caption{The learning rate and batch size that resulted in the best classification accuracy for each model.}
\label{table:hyperparameters}
\end{table}

\end{document}